%% file: sample-sigconf.tex
\documentclass[sigconf]{acmart}

\usepackage{booktabs} 

\usepackage{multirow}
\usepackage[export]{adjustbox}
\usepackage{booktabs} 
\usepackage{multirow}
\usepackage{color}
\usepackage{graphicx}
\usepackage{mathtools}
\usepackage{amsmath}

\DeclareMathOperator*{\argmax}{arg\,max}

\setcopyright{rightsretained}

\acmDOI{10.475/123_4}

\acmISBN{123-4567-24-567/08/06}

\begin{document}

\title{Semantic Summarization of Egocentric Photo Stream Events}

\author{Aniol Lidon}
\affiliation{%
  \institution{Universitat Politecnica de Catalunya}
  \city{Barcelona} 
  \state{Catalonia/Spain} 
}

\author{Marc Bola\~{n}os}
\affiliation{%
  \institution{Universitat de Barcelona}
  \city{Barcelona} 
  \country{Spain}}
\email{marc.bolanos@ub.edu}

\author{Mariella~Dimiccoli}
\affiliation{%
  \institution{Universitat de Barcelona}
  \institution{Computer Vision Center}
  \city{Barcelona}
  \country{Spain}
  }
\email{mariella.dimiccolig@cvc.uab.es}

\author{Petia Radeva}
\affiliation{%
  \institution{Universitat de Barcelona}
  \city{Barcelona} 
  \country{Spain}}
\email{petia.radeva@ub.edu}

\author{Maite Garolera}
\affiliation{%
  \institution{Consorci Sanitari de Terrassa}
  \city{Terrassa}
  \country{Spain}
  }
\email{maite.garolera@cst.cat}

\author{Xavier Giro-i-Nieto}
\affiliation{%
  \institution{Universitat Politecnica de~Catalunya}
  \city{Barcelona} 
  \country{Catalonia/Spain} 
}
\email{xavier.giro@upc.edu}

\renewcommand{\shortauthors}{A. Lidon et al.}

\begin{abstract}
\input{0_abstract} 
\end{abstract}

\copyrightyear{2017} 
\acmYear{2017} 
\setcopyright{acmcopyright}
\acmConference{LTA'17}{October 23, 2017}{Mountain View, CA, USA}\acmPrice{15.00}\acmDOI{10.1145/3133202.3133204}
\acmISBN{978-1-4503-5503-2/17/10}

%
%





\maketitle

\input{1_introduction}

\input{scheme}
\input{2_related_work}

\input{3_methodology}
\input{3_2_informativeness}

\input{3_3_diversity}

\input{3_4_novelty}

\input{4_setup}

\input{5_results}
\input{6_conclusions}
\input{7_acks}

\bibliographystyle{ACM-Reference-Format}
\bibliography{references} 

\end{document}

%% file: 0_abstract.tex
With the rapid increase of users of wearable cameras in recent years and of the amount of data they produce, there is a strong need for automatic retrieval and summarization techniques. This work addresses the problem of automatically summarizing egocentric photo streams captured through a wearable camera by taking an image retrieval perspective. After removing non-informative images by a new CNN-based filter,  images are ranked by relevance to ensure semantic diversity and  finally re-ranked by a novelty criterion to reduce redundancy.  To assess the results, a new evaluation metric is proposed which takes into account the non-uniqueness of the solution. Experimental results applied on a database of 7,110 images from 6 different subjects and evaluated by experts gave 95.74\% of experts satisfaction and a Mean Opinion Score of 4.57 out of 5.0.
Source code to reproduce this work is available at \url{https://github.com/imatge-upc/egocentric-2017-lta}.

%% file: 1_introduction.tex
\section{Introduction}

From smartphones to wearable devices, digital cameras are becoming ubiquitous. This process is being accompanied by the progressive reduction of digital storage cost, making it possible to collect large amounts of high-quality pictures in a easy and affordable way.
This situation arises a number of natural questions: how to manage this large amount of pictures? Do we really need to store all of them? Summarization, the process of generating a proper, compact and meaningful representation of a given image collection through a subset of representative images, is crucial to help managing and browsing efficiently large volumes of video content. Although summarization is not a new research topic in Computer Vision \cite{rajendra2014survey,gygli2014creating}, it is still a largely open problem. 

Our goal in this paper is to address the summarization problem focusing on a particular scenario: the one of analyzing photo streams acquired through a wearable camera. 
Motivation for this work is given by the explosion of the number of wearable camera users in recent years and, consequently, by the huge amount of data they produce. To record daily experiences from an egocentric, first-person perspective is a trend that has been growing progressively since 1998, when Steve Mann proposed the WearCam \cite{wearcam}. In 2000, Mayol et al. proposed a necklace-like lifelogging device \cite{Mayol} and, in 2006, Microsoft Research started to commercialize the first egocentric lifelogging portable camera, the SenseCam, for research purposes \cite{SenseCam}.
 
  \begin{figure}[!t]	\includegraphics[width=\columnwidth]{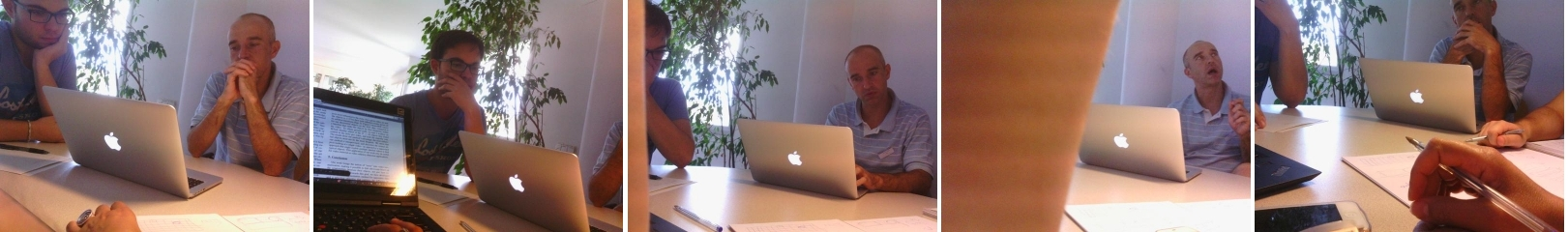} 
    \caption{Example of temporal neighbouring images acquired by a wearable photo camera.}
    \label{fig:example}
\end{figure}
 

Several authors \cite{Sellen, Lee, Piasek2011, piasek2014using} have studied the benefits of lifelogging cues such as egocentric images to help people with dementia to enhance their memory or to help them remember about their forgotten past. Sellen et al. \cite{Sellen} showed  that  episodic  details  from a visual 'lifelog' can be presented to users as memory cues to assist them in remembering the details of their original experience. To support people with dementia, Piasek et al. \cite{Piasek2011} introduced the "SenseCam Therapy" as a therapeutic approach similar to the well established "Cognitive Stimulation Theraphy" \cite{CogST}. Participants were asked to wear SenseCam in order to collect images of events from their everyday lives,  then images were reviewed with a trained therapist. 


However, lifelogging technologies produce huge  amounts of data (thousands of images per day) that should be reviewed by both patients and caregivers. To be efficient, lifelogging systems need to summarize the most relevant information in the images. On the other hand, in order to make possible the recording of images from the whole day, it is necessary to use wearable cameras with low temporal resolution (2 fpm). The peculiarity of these image collections is that, due to the low-temporal resolution of the camera and to the free motion of its wearer, temporally adjacent images may be very different in appearance even if they belong to the same event and should therefore be grouped together. For instance, during a meeting, the people the camera wearer is interacting with and the objects around them may change their position and frequently appear occluded (see Fig. \ref{fig:example}).

As a consequence of this, taking the semantics into account is crucial to summarize egocentric sequences acquired by a low temporal resolution camera. This paper proposes a method to summarize each of the events present in a daily egocentric sequence, aiming at preserving semantic information and diversity, while reducing the total number of images. The method consists of three major steps: first, non-informative images are removed; second, they are ranked by semantic relevance; and finally, a new re-rank is applied by enforcing diversity among the chosen subset of pictures. 
Our contributions can be summarized as follows:
\begin{enumerate}
\item Propose a CNN-based informativeness estimator for egocentric images.
\item Define a set of semantic relevance criteria for egocentric images.
\item Formulate the summarization task as a retrieval problem by combining informativeness, relevance and novelty criteria.
\item Define a soft metric to assess the novelty from partially annotated image datasets.
\item Our results have been validated by medical experts  with the aim of being used in a cognitive training framework to reinforce the memory of patients with mild cognitive impairment.
\end{enumerate}

The rest of the paper is organized as follows: the next section reviews the related work, section \ref{sec:methodology} details the proposed method, section \ref{sec:setup} describes our experimental setup, section \ref{sec:results} presents the experimental results and finally, section \ref{sec:conclusions} ends the paper with some concluding remarks. 
Source code to reproduce this work is available at \url{https://github.com/imatge-upc/egocentric-2017-lta}.

%% file: scheme.tex
\begin{figure*}[!ht]
	\begin{center}
	\includegraphics[width=\textwidth]{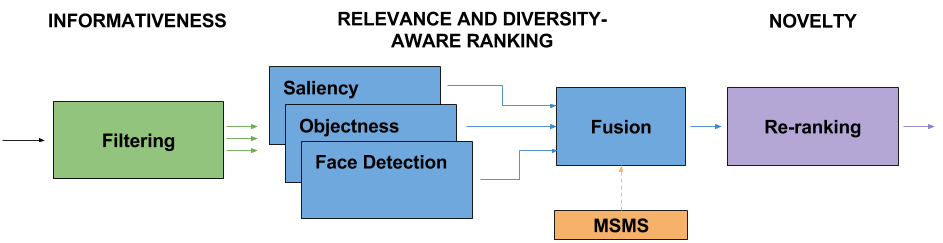} 
    \caption{General scheme of our  event summarization methodology.}
    \label{fig:scheme}
    \end{center}
\end{figure*}

%% file: 2_related_work.tex
\section{Related Work}


\subsection{Summarization by key-frame selection in Lifelogging}

Egocentric photo stream summarization has been traditionally formulated as the problem of grouping lifelog images into coherent collections (or events) by: first, extracting low-level spatio-temporal features and then, selecting the most representative image from each event. In this spirit, many authors proposed different strategies for temporal segmentation and key-frame selection. 
Doherty et al. \cite{Doherty} proposed a key-frame selection technique, which seeks to select the image with the highest 'quality' as key-frame by relying on five types of features: contrast, color variance, global sharpness, noise, saliency and external sensors data (accelerometers and light).
In addition to image quality, Blighe et al. \cite{Blighe} considered image similarity to select a key-frame in an event. Basically, after applying an image quality filter that removes all poor quality images, they selected the image which has the highest average similarity to all other
images in the event as the key-frame. In that case, similarity was measured relying on the distance of SIFT descriptors.
In \cite{Jinda}, the key-frame is selected by a nearest
neighbour ratio strategy that favors high quality images and, if the difference in quality is not large enough, favors images closer to the middle of the temporal segment. More recently, the auhors in   \cite{bolanos2015visual} proposed to use a Random Walk for selecting a single and most representative image for each temporal segment.

While these methods rely solely on low-level or mid-level features for the temporal segmentation and key-frame detection, a few recent works have introduced a higher
semantic level in the selection process for video cameras. Although, in this cases, due to the higher temporal resolution of the camera (about 30fps), aiming at selecting subshots (short video sub-sequences) instead of unique key-frames.
Lu and Grauman \cite{lu2013story} and lately Ghosh et al. \cite{ghosh2012discovering} suggested that video summarization should preserve the narrative character of a visual lifelog and, therefore, it should ideally be made of a coherent chain of video subshots in which each subshot influences the next through some subset of key visual objects. Following this idea, in \cite{lu2013story,ghosh2012discovering}, first important people and objects are discovered based on their interaction time with the camera wearer and then, a subshot selection driven by key-object event occurrences is applied. Subshot selection is performed by incorporating into an objective function a term corresponding to the influence between subshots as well as image diversity. To model diversity, the authors relied on GIST descriptors and color histograms to model scenes and proposed a measure of diversity that is high when the scenes in sequential subshots are dissimilar to ensure visual uniqueness. Visual diversity in lifelog summaries is modeled in \cite{aghazadeh2011novelty} through the concept of novelty, which the authors heuristically defined as the deviation from some standard background. According to this definition, novelty is detected based on the absence of a good registration, in terms of ego-motion and the environment,  between a new sequence and stored reference sequences.
More recently, Gong et al. \cite{NIPS2014_5413} proposed the so called Sequential Determinantal Point Process (seqDPP) approach for video summarization, a probabilistic
model with the ability to teach the system how to select informative and diverse subsets from human-created summaries, so as to best fit human-perception quality based on evaluation metrics. This work is an adaptation to sequences of \cite{kulesza2012determinantal}, which is able to capture the strong dependency structures between items in sequences.
It is worth to mention that 
all these works have been conceived to deal with video data, where, assuming that the temporal segmentation is good, temporal coherence and frame redundancy make the key-frame selection process easier. 

\subsection{Diversity and novelty in information retrieval}

One of the first works that tried to approach the problem of obtaining a diverse set of elements was presented in 1998 by Carbonell \& Goldstain \cite{Carbonell}. Their proposal, which was applied to the context of text retrieval and summarization, aimed at obtaining results highly relevant for the query, but presenting a low redundancy. They define the marginal relevance as the linear combination of relevance and novelty, measured independently. They aimed at maximizing it iteratively, defining this way what they call the Maximal Marginal Relevance (MMR).

A similar formulation to MMR of the diversification problem was given more recently by Deselaers et al. \cite{Deselaers} in the problem of image retrieval. Likewise, they jointly optimized the relevance and the diversity of the query, although reformulating the general solution by incorporating dynamic programming techniques to the initial proposal.


Diversity in social image retrieval was one of the focus of the MediaEval 2013, 2014 and 2015 benchmarks and attracted the interest of many groups working in this area. Most participants developed diversification approaches that  combined clustering with a key-frame selection strategy to extract representative images for each cluster. Spyromitros et al. proposed an  MMR-based approach \cite{spyromitros} that jointly considers relevance and diversity, but using a supervised classification model to obtain the relevance scores learned from the user feedback.
The contribution from Dang-Nguyen et al. \cite{Dang-Nguyen} was to filter out non-relevant images at the beginning  of the process before applying diversity, hence simplifying it.

In \cite{Song}, the authors presented a method for detecting and resolving the ambiguity of a query based on the textual features of the image collection. If a query has an ambiguous nature, this ambiguity should be reflected in the diversity of the result to increase user satisfaction. 
Leuken et al. \cite{Leuken} proposed to reduce the ambiguity of results provided by image search engines relying on textual descriptions by seeking for the visual diversification of image search results. This is achieved by clustering the retrieved images based on their visual similarity and by selecting a representative image for each cluster.
In these works 'diversity' is aimed at addressing the 'ambiguity' of the textual descriptions (tags) in queries rather than avoiding redundancy in search results. The same terminology is used by \cite{clarke2008novelty}, where the term 'novelty' is meant to address redundancy in the retrieved documents. In this work we will use the terms novelty and diversity as defined in \cite{clarke2008novelty}.

%% file: 3_methodology.tex
\section{Methodology}
\label{sec:methodology}
Inspired by the image retrieval work, we question which would be the most suitable diversity and relevance criteria applied on egocentric images that kept the narrative character of the visual lifelog. Taking into account that the images are acquired non-intentionally, it is important to disregard the non-informative images (see Section \ref{sec:informativeness}) and keep the minimal set that represents the visual event.
In this section, we explain 
the four main steps applied to construct the final resulting summary (see Fig. \ref{fig:scheme}): 

\textbf{(1) Informativeness Filtering} (Section \ref{sec:informativeness}): egocentric images are acquired non-intentionally and therefore, many of them may be non-informative, capturing neither (or partially) objects nor people, or being blurred or dark. By means of a CNN-based  informativeness filtering method, we discard  most of the non-informative images from the egocentric event.

\textbf{(2) Relevance and Diversity-aware Ranking} (Section \ref{sec:relevance}): an initial relevance image ranking is computed taking into account different criteria such as Saliency and Objectness for dealing with the ambiguity or under-specification of queries like \emph{What is the user doing?}, or \emph{Where is the user?}, and Face Detection for answering the question \emph{With whom is the user most likely interacting?}. 

\textbf{(3) Novelty-based Re-ranking} (Section \ref{sec:novelty}): a final re-ranking based on a novelty-maximization procedure is applied on the images already ranked by relevance. This step is crucial to select those images that represent the most varied set of concepts appearing avoiding redundancy without semantic loss.

\textbf{(4) Estimation of Fusion Weights with Mean Sum of Maximal Similarities} (Section \ref{sec:metrica}): in this step, we define a novel soft metric, which we call Mean Sum of Maximal Similarities (MSMS), in order to define the priorities of the different relevance terms and construct the final summary. 

%% file: 3_2_informativeness.tex
\subsection{Informativeness Filtering} \label{sec:informativeness}

\begin{figure}[!t]
	\begin{center}
	\includegraphics[width=\columnwidth]{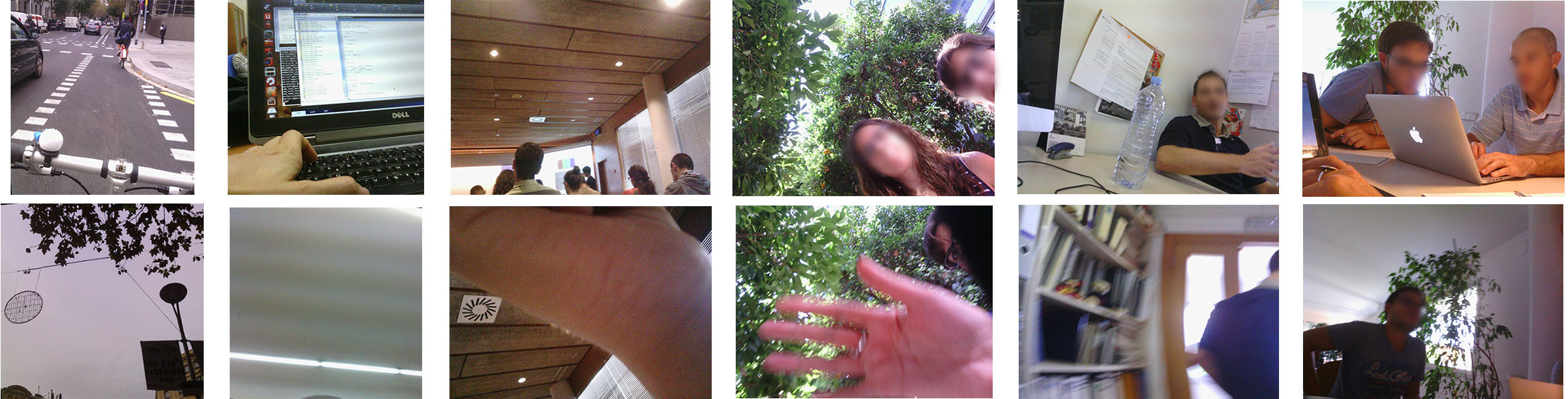}
    \caption{Examples of informative images (top) and non-informative images (bottom) belonging to the same events. Faces appearing in the images have been manually blurred for privacy concerns.}
	\label{fig:informative_examples}
    \end{center}
\end{figure}

Considering both the free motion of egocentric cameras and the non-intentionality of the pictures they take, several problems are inherent to them as over- or under-light exposure, blurriness, pictures of the sky or ground, pictures where possible objects of interest are badly centered in the image or even non-existent, etc. Such images are considered as {\em non-informative pictures}.
A  way to avoid a good amount of the undesired images in the final summary and also to boost the performance of the next steps in our methodology, would be being able to discriminate  the {\em informative pictures}. Fig. \ref{fig:informative_examples} illustrates informative vs. non-informative photos taken by the wearable camera.

In order to learn a model of  informative pictures, and taking into account the complexity of distinguishing visually whether an image is informative enough or not, we propose training a CNN for a binary problem. Thus, all the images with an undesired artifact (empty image, blurred image, image with small amount of information, image where something occludes most of the image region, sky, ground, ceiling, room wall, etc.) will be considered non-informative (label 0), and the rest (images with semantic content) will be considered informative (label 1). With this procedure, we will be able to extract an {\em informativeness score} for each image and filter the unuseful ones. 
In order to remove as many images as possible, but with the ultimate goal of having a very high recall (always try to keep any informative image for the next steps), we only discard the images with $informativenessScore < 0.025$, which correspond to the ones that are considered
non-informative for certain by the CNN (see experimental results in section \ref{sec:informativeness_results}).

The network training for the binary class distinction is performed by fine-tuning the CaffeNet \cite{krizhevsky2012imagenet} pre-trained on the ImageNet \cite{deng2009imagenet} dataset, provided in the software Caffe \cite{jia2014caffe}.

%% file: 3_3_diversity.tex
\begin{figure}[t]
\centering
\includegraphics[width=3.5in]{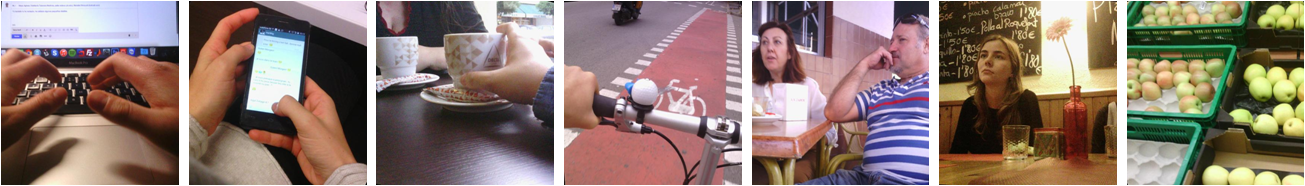}
\caption{Images acquired by a lifelogging device, where objects of interest appear like: computer, mobile, coffee, hand, bicycle, person, face, flower, apple, etc.}
\label{fig:lifelog1}
\end{figure}

\subsection{Relevance and Diversity-aware
Ranking}
\label{sec:relevance}
Once non-informative images are discarded, we proceed to rank the remaining images by considering a relevance criteria.
Our solution formulates the summarization problem in similar terms as in information retrieval.
A ranked list of event frames is generated in such a way that a summary of $T$ images directly corresponds to a truncation of the ranked list of $N$ elements at its $T$-th position.
Classic retrieval problems build their ranked lists as a response to a user query, in the summarization of events, this query would correspond to: \textit{Select $T$ images to describe the event depicted into these $N$ frames.}
This query is highly ambiguous as an event can be described from different perspectives.
For example, an egocentric visual summary of an event may respond to multiple intentions, such as: 
\textit{Where} is the user? \textit{What} activity is the user performing? \textit{With whom} is the user interacting? 
We hypothesize that the relevance of each image with respect to these questions can be estimated with computer vision tools for saliency prediction, detection of objects and detection of faces.


\subsubsection{Saliency Prediction for image relevance}
We assume that images with more salient content are relevant and should have a higher probability to be included in the summary.
Visual saliency can be triggered by a broad range of reasons, such as objects, people or characteristic features appearing in the picture. Many computer vision algorithms try to estimate the fixation points of the human eyes in a scene by means of \textit{saliency maps}. These are heat maps of the same size of the image, whose higher values correspond to the image locations with a higher probability of capturing the human visual attention.

In this work, we compute the saliency maps with SalNet \cite{pan2015end}, an end-to-end convolutional network for saliency prediction. We adopt the overall sum of the values in the saliency map as a quantitative estimator of the image relevance.

\subsubsection{Object Detection for image relevance}
We assume that those images containing objects are relevant since these objects likely correspond to the ones the user is interacting with.
These objects would address the question of \textit{What is the user doing?} (see Fig. \ref{fig:lifelog1}).
In addition, introducing a semantic interpretation of the scene, targets the summary from a higher abstraction level than saliency maps.


In our work, we used the off-the-shelf tool \textit{Large Scale Detection through Adaptation (LSDA)} presented in \cite{lsda}.
This object detector is based on a CNN fine-tuned for local scale and provides a semantic label and a confidence score in the localization of the objects in the scene.

This tool allows to estimate the relevance of each frame by summing the detection scores of all objects in the picture, so that the frames with higher confidence detection will be considered as more relevant.


\subsubsection{Face Detection for image relevance}

We assume that images containing people are also relevant, since these people likely correspond to the ones the user is interacting with, and they would be useful to answer the question \textit{With whom is the user interacting?}.
Therefore, a face detector complements the object detector into providing a cue for the user social interactions during the event.



In our solution, we adopted the off-the-shelf  face detector by Zhu et al. \cite{zhu2012face}, which provides a confidence for each of the detected faces.
The relevance of each frame in terms of social interaction was estimated  by summing the confidence scores of the face detectors. In the particular implementation of \cite{zhu2012face}, detection scores may also be described with negative values, so we actually used these scores in an exponential sum, which conveniently deals with the negative scores as well as encourages the selection of frames with multiple detected faces.

\subsubsection{Diversity-aware Ranking}
\label{sec:diversity}

The three criteria used for the relevance detailed above allow to cope with the ambiguity of the query, since they estimate the relevance from three different perspectives. The relevance scores computed for each case are then used to build three ranked lists that will be combined into a single one. 
This combination is based on generating a set of normalized scores based simply on the position of the frames. 
Normalized scores $r_k(x)$ are linearly distributed from the top (1 for most relevant) to the bottom (0 for non relevant) of the ranked list as follows:
\begin{equation*}
r_k(x) = \frac{M - R_k(x)}{M - 1},
\end{equation*}
where $R_k(x)=1,\ldots,M$ is the ranking position associated with image $x$ according to each relevance criterion $k \in \{1,2,3\}$ and $M$ is the number of informative frames in the event, being $M<N$ and N the total number of images in the event.
The standard score normalization \cite{montague2001relevance, renda2003web} with the min and max scores was also tested giving similar results, so the rank-based normalization was adopted to save computational resources.

%% file: 3_4_novelty.tex
\subsection{Novelty-based Re-ranking} 
\label{sec:novelty}
The relevance and diversity-aware ranked list sorts the informative images combining different criteria for relevance, but does not explicitly cope with the redundancy in the information. 
Further processing is necessary to maximize the \textit{novelty} provided by each image with respect to the rest in the summary.
In information retrieval, \textit{novelty} is defined as a quality of a system that avoids redundancy \cite{clarke2008novelty}.
In our approach for building a summary by truncating a ranked list of images, introducing novelty implies that each image in the list should differ as much as possible from its predecessors.


Our approach adopts the \textit{greedy} selection algorithm presented by Deselaers et. al \cite{deselaers2009jointly} to re-rank the fused list based on novelty.
The goal of our summarization algorithm 
is to analyze the input set of informative and ranked images $\mathcal{X}=\{x_{1}, ...x_{M}\}$ to iteratively build another set  with minimal redundancy, say $\mathcal{Y}^T=\{x_{y_1}, ...x_{y_T} \}$, where $T \leq M$.
Our approach starts by selecting the top ranked image $x_1$ in the diversity-aware ranked list as the first element of $\mathcal{Y}$, that is $\mathcal{Y}^1=\{ x_{y_1} \}$.
The \textit{novelty} of each candidate image, $x^*$ to be added to the summary at iteration $t = 2,...,T$ is defined as:

\begin{equation}
\label{eq:novelty}
n(x^*, \mathcal{Y}^t)= 
1 - s(x^*,\mathcal{Y}^t)=
1 - \smash{\displaystyle\max_{x_{y_j} \in \mathcal{Y}^t}} s(x^*,x_{y_j})
\end{equation}
 $s(x^*,x_{y_j})$ is a normalized similarity measure between the candidate $x^*$ and image $x_{y_j}$.
In this way, the more different a new image is with respect to the ones in $\mathcal{Y}^t$, the higher its novelty is.

In our work, the similarity $s(x^*,x_{y_j})$ is based on visual appearance.
Each image is described by a feature vector corresponding to the seventh fully-connected layer of the CaffeNet convolutional neural network \cite{jia2014caffe}, which was also used as initialization for the informativeness (see section \ref{sec:informativeness}), and whose architecture was inspired by AlexNet \cite{krizhevsky2012imagenet} and trained with the ImageNet dataset \cite{deng2009imagenet}.
The similarity score is computed with the Euclidean distance of the feature vectors.

The summary of the event is built by combining the relevance and the novelty of the candidate frames, $x^*$.  
A greedy selection algorithm iteratively chooses the next image in the summary as:

\begin{equation}
\label{eq:greedy}
\begin{array}{rl}
 x_{y_{t+1}} = & \smash{\displaystyle \argmax_{x^* \in \mathcal{X} \setminus \mathcal{Y}^t}} \
(r(x^*) + n(x^*,\mathcal{Y}^t)),
\\
\\
\vspace*{2mm}
\mathcal{Y}^{t+1}= & \mathcal{Y}^t \cup \{ x_{y_{t+1} } \}
\end{array}
\end{equation}
that is, the image with the highest sum of relevance and novelty. We detail the computation of $r(x^*)$ in Equation (\ref{eq:fusion}). 




\subsection{Estimation of Fusion Weights with Mean Sum of Maximal Similarities}\label{sec:metrica}

\begin{figure}[t]
\centering
\includegraphics[width=\columnwidth]{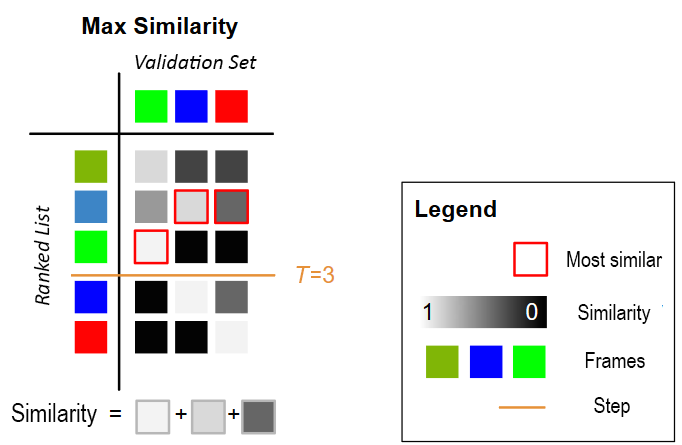}
\caption{Visual representation of the Sum of Max Similarities when comparing the first $T=3$ images from the ranked list with the $P=3$ red, green and blue images from the validation set. The similarity between images from the validation set  and images from the ranked list  is represented as a $T \times P$ matrix of gray level squares, where each square $(i,j)$ of the matrix represents the similarity between the image $i$ from the ranked list and the image $j$ from the validation set. High intensity gray level values indicate high similarity.}
\label{fig:hung_sms}
\end{figure}


Different relevance criteria would have different importance for the final ranking. 
To this aim, we propose a novel approach to estimate the priorities of each criterion and fuse them, based on comparing the similarity of the images from a validation set of $P$ elements  $\mathcal{V}=\{x_{v_1},...,x_{v_P}\}$ with a summary of $T$ elements $\mathcal{Y}^T= \{x_{y_1}, ...x_{y_T} \}$.

The \textit{Sum of Maximal Similarities (SMS)} of $\mathcal{V}$ with respect to $\mathcal{Y}^T$ is defined as:
\begin{equation}
\label{eq:noveltysum1}
SMS(\mathcal{V}, \mathcal{Y}^T) =
\frac{1}{P}
\sum_{i=1}^{P} s(x_{v_i}, \mathcal{Y}^T)
\end{equation}
where $s(x_{v_i}, \mathcal{Y}^T)$ is the similarity of image $x_{v_i}$ from the validation set with respect to $\mathcal{Y}^T$. 
Following this metric, the more similar is the validation set $\mathcal{V}$ to the selected images $\mathcal{Y}^T$, the highest the average similarity. 

The main advantage of our soft metric presented in Equation (\ref{eq:noveltysum1}) is that automatic summaries, 
which are very similar to the validation images, although do not coincide, still will be assessed as a valid solution. 
This feature of our soft metric is specially important when working with sequences of egocentric photo streams, that usually contain a high amount of redundancy.

Our summaries are built as a truncation of a ranked list, so the value of $SMS$ depends on $T$.
Fig. \ref{fig:hung_sms} shows a schematic example, where a validation set composed of $P=3$ images (represented by three colors: green, blue and red) is compared with an event of $M=5$ images, which have been previously filtered and ranked based on our diversity and novelty-aware criterion.
In this example, a summary with $T=3$ images is considered. Let us imagine that the green image from $\mathcal{V}$ is matched with the third one in the ranked list, while the blue and red images are matched with the second one.

Notice that, as in our set up the validation set $\mathcal{V}$ is always a subset of the input set $\mathcal{X}$, hence, the final average similarity (i.e. considering the whole sequence, $T=M$ images)  for $\mathcal{Y}^M$ will always correspond to one. Actually, in the example of Fig. \ref{fig:hung_sms}, SMS must reach a value of one when all the ranked list is considered, that is, when $T=M$.

 \begin{figure}[t]
\centering
\includegraphics[width=0.8\columnwidth]{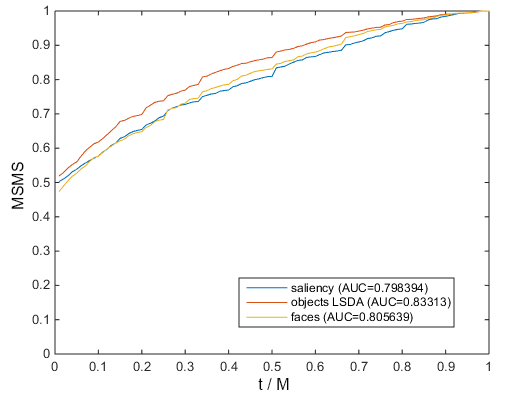}
\caption{MSMS curves and AUC for each relevance criterion used separately.}
\label{fig:mnsms}
\end{figure}

Let us consider the evolution of the $\overline{SMS}$ of summary $\mathcal{Y}$ as the parameter $t$ grows ($t=1,2,\ldots, M$) 
defined as:
\begin{equation*}
\label{eq:noveltysum2}
\overline{SMS}(\mathcal{V}, \mathcal{Y}) = \{ 
 SMS(\mathcal{V}, \mathcal{Y}^1),\ldots,  
\\
SMS(\mathcal{V}, \mathcal{Y}^M) \}
\end{equation*}

We construct the $\overline{SMS}$ curves for all the validation set, interpolate them, normalize them with respect to the length $M$ of each sequence and get the average of the curves, one curve for each event in the validation set. The resulting curve represents the \textit{Mean Sum of Maximal Similarities (MSMS)}.




Fig. \ref{fig:mnsms} shows three examples of MSMS curves. The  curve illustrates the evolution of the MSMS as a function of the percentage of images covered by the visual summary.
This curve is defined over the $X$-axis representing the proportion of event images represented in the top $t$ items in the list, that is, plotting the $MSMS$ over $t/M$, where $M$ is the amount of already filtered images in the event. 
In this way, the curves of events of different lengths $M$ can be compared on the same plot.
The best MSMS curves are those which reach higher values with the minimal amount of images in the summary.



Note that the MSMS can be computed for the three relevance criteria that can give as a quality of performance of each of them. Hence, we introduce a final refinement in the fusion of the ranked lists by estimating the confidence of each of the three relevance criteria and using it to weight the corresponding relevance terms.
Thus, we leverage the contribution of each of the three relevance criteria based on their stand-alone performance
(see Fig. \ref{fig:mnsms}).
This weight corresponds to the normalized Area Under the Curve (AUC) of the MSMS measure.

As a result, the fusion weight $w(k)$ for the relevance criterion $k$ is estimated as:
\begin{equation*}
w(k) = \frac{AUC(k)}{\sum_{i=1}^3 AUC(i)}.
\label{eq:weight}
\end{equation*}

Finally, the three normalized relevance scores associated to each frame are aggregated with a weighted sum to obtain their fused score $r(x)$, as described by the following equation:

\begin{equation}
\label{eq:fusion}
r(x) = \sum_{k=1}^3 w(k) \ r_k(x),
\end{equation}
and the final set of frames is obtained according to the updated relevance and novelty criteria according to Equation (\ref{eq:greedy}).

%% file: 4_setup.tex
\section{Experimental setup}
\label{sec:setup}

\subsection{Dataset} 
\label{ssec:dataset}

The dataset used for validation of our method was acquired with the wearable camera Narrative Clip (www.getnarrative.com), which takes a picture every 30 seconds (2 fpm). It is composed of 10 day lifelogs from 5 different subjects, with a total of 7,110 images\footnote{The link to the dataset and its ground truth are prepared to be made public domain when the article is published.}. Each day has been segmented in between 10 to 25 events manually, although any automatic segmentation method can be used (e.g. \cite{talavera2015r}). The event segmentation separated the pictures in a set of ordered and relevant semantic events or segments. In this paper, we use this segmentation as starting point for the semantic summarization. 


In order to apply a quantitative evaluation of our method, each day lifelog was annotated at two levels by psychologists in the following way: 


\textbf{(1) GT Level 1 - Informativeness:} positive or negative label depending on whether the image is considered informative or non-informative (see section \ref{sec:informativeness}). The proportion of images labeled as informative is $61.22\%$ of the complete dataset.

\textbf{GT Level 2 - Grouping of similar images:} all highly similar informative images that belong to the same event are grouped together (see section \ref{sec:diversity}). This distinction, resembling the one used in \textit{MediaEval 2014 Retrieving diverse social images challenge} \cite{ionescu2014retrieving}, intends to provide a way of measuring how many different clusters/groups from the ground truth are represented among the results in the final diversity selection.


\subsection{Semantic Assessment of Summaries} 
\label{ssec:metrics}

The assessment of visual summaries is a challenging task due to the rich semantic content of the images and the ambiguity in the evaluation criteria. 
In addition, human annotation is an expensive resource, which becomes dramatically scarce, when dealing with tasks that require expert annotations in the domain. In our case, the ground truth annotations have been performed by psychologists who defined the event's summaries 
from an human-centered point of view.


We addressed the challenge of summary assessment as follows: 
starting from the daily events defined by the psychologists,
the automatic summaries were built for each of the events separately by using the best configuration of our system. Later, the results were presented in a blind taste test to the expert annotators. 



\subsubsection{Validation based on MSMS}

The classical methods to evaluate summaries cannot be applied in our setup, because they require the annotation of the full dataset.
In a traditional case, each document in the ground truth is labeled as relevant or non-relevant for the summary and, in some cases, also clustered in groups of redundant items that cover the same sub-topic.
Such annotations allow the definition of metrics like \textit{Precision} \cite{Carbonell} for relevance and \textit{Cluster Recall (or subtopic-recall)} \cite{zhai2003beyond} for diversity and/or novelty. 
However, in other cases (like ours), only a small portion of the dataset is annotated.

Given this limitation, following the framework of Subsection \ref{sec:novelty}, we propose an evaluation approach based on the similarity of the images in the ground truth set $\mathcal{G}=\{x_{g_1},...,x_{g_P}\}$, when compared to the automatic summary of $T$ elements, $\mathcal{Y}^T=\{x_{y_1}, ...x_{y_T}\}$.
In this case, we apply the SMS, $ SMS(\mathcal{G}, \mathcal{Y}^T)$ of ground truth $\mathcal{G}$ with respect to the extracted summary, $\mathcal{Y}^T$.
As before, we obtain the MSMS as the average value, when considering all events in the dataset of the ground truth, 
which is finally plotted in the $Y$-axis of our evaluation scores. 
Finally, the AUC is computed to obtain a quantitative measure of each configuration for the summary.

\subsubsection{Blind taste test}
\label{ssec:expert_eval}


The resulting summaries are evaluated with a \textit{blind taste test} by the team of experts who generated the ground truth summaries. 
This methodology, previously used in \cite{lu2013story,bolanos2015visual}, shows different summaries from the same event to the experts, so they can rate them from a comparative perspective.
By randomly changing the position of the different techniques at each event, graders are unaware of what technique was used to generate each of the summaries, guaranteeing this way that their judgments are not biased towards any of the algorithm components.




The amount of images included in each summary corresponds to the number of frames selected by the experts, when building the ground truth summaries, that is, $P = T$.
In this way, the summary shown to the graders is a truncation of the final ranked list of images. In our case, the experts decided the length of the automatic summary to be a percentage of the length of the whole event. 
Note that developing an automatic system to establish the length of the summary is out of the scope of this paper.



Additionally, the selected images are sorted in the summary according to their temporal time stamps.
This final sorting helps the user to reconstruct the story between the images to have a better understanding of the event.
An online platform was developed for the experts to evaluate our results. In a first round of questionnaires, we wanted to compare the result of our diversity method using ImageNet or Places as a similarity criterion (see Equation (\ref{eq:novelty})).
For each event, the questionnaire first shows all the images belonging to the event, and then the user has to answer the questions \textit{Is the summary representative of the event?} and \textit{Which summary do you prefer?} for each similarity method (using ImageNet or Places).


On the second questionnaire the expert had to give a grade to each of the presented ranked summaries.
The three summaries correspond respectively to: our approach, a uniform sampling of images as a lower-bound example, and the ground truth summaries constructed by the same experts two months earlier as an upper-bound of the performance score.
The order in which the different summaries are shown is also randomly chosen for each event to avoid again any bias due to the sorting.


In this case, a comparison between the three types of summaries is obtained by asking the experts to grade each solution from 1 (worst) to 5 (best).
This data collected allowed the computation of the \textit{Mean Opinion Score (MOS)} for each configuration (see section \ref{subsection:summary_quality}).




%% file: 5_results.tex
\section{Results}
\label{sec:results}
Two types of experiments validate the presented results: one for the informativeness filtering and a second one using online questionnaires for the relevance, diversity and novelty on the final visual summaries. In both cases, the evaluation was based on the feedback provided by the experts.\\

\subsection{Informativeness Filtering}\label{sec:informativeness_results}

\input{table_informativeness}

\begin{figure}[!h]
	\begin{center}
	\includegraphics[width=0.75\columnwidth]{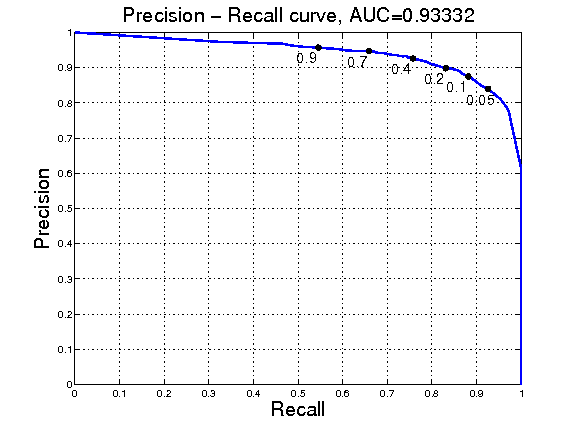} 
    \caption{Precision-Recall curve for different informativeness score threshold values (a small subset of them are shown in black dots) for all the sets.}
    \label{fig:prec_recall_informativeness}
    \end{center}
\end{figure}

To validate how successful is our algorithm to filter non-informative frames, we applied a 10-fold (a day set out) cross-validation. For all the experiments, we used the following parameters: $base\_lr = 10^{-8}$, $lr\_policy = "step"$, $gamma = 0.1$, $stepsize = 3000$,  $blobs\_lr_{fc6} = blobs\_lr_{conv5} \times 5$, $blobs\_lr_{fc7} = blobs\_lr_{conv5} \times 8$ and $blobs\_lr_{fc8} = blobs\_lr_{conv5} \times 10$. In Table \ref{tab:cross_val_informativeness}, we present the validation accuracy obtained on each of the sets and the number of iterations applied to achieve the best result. 
After having the networks trained for each cross-validation, we evaluated the general average performance in terms of Accuracy, Precision, Recall and F-Measure for different informativeness score threshold values\footnote{If the informativeness score of a sample is below the threshold, it will be considered non-informative.}. As we can see, we are able to obtain the highest F-Measure value, when we filter images with an $informativenessScore < 0.025$. In Fig. \ref{fig:prec_recall_informativeness}, we show the Precision-Recall curves for all thresholds, obtaining the best results with respect to the F-Measure of: $FMeasure = 0.881$, $Accuracy = 0.846$, $Precision = 0.84$ and $Recall = 0.926$ with $threshold = 0.05$.\\


\subsection{Diversity}

\begin{figure}[!h]
	\begin{center}
	\includegraphics[width=\columnwidth]{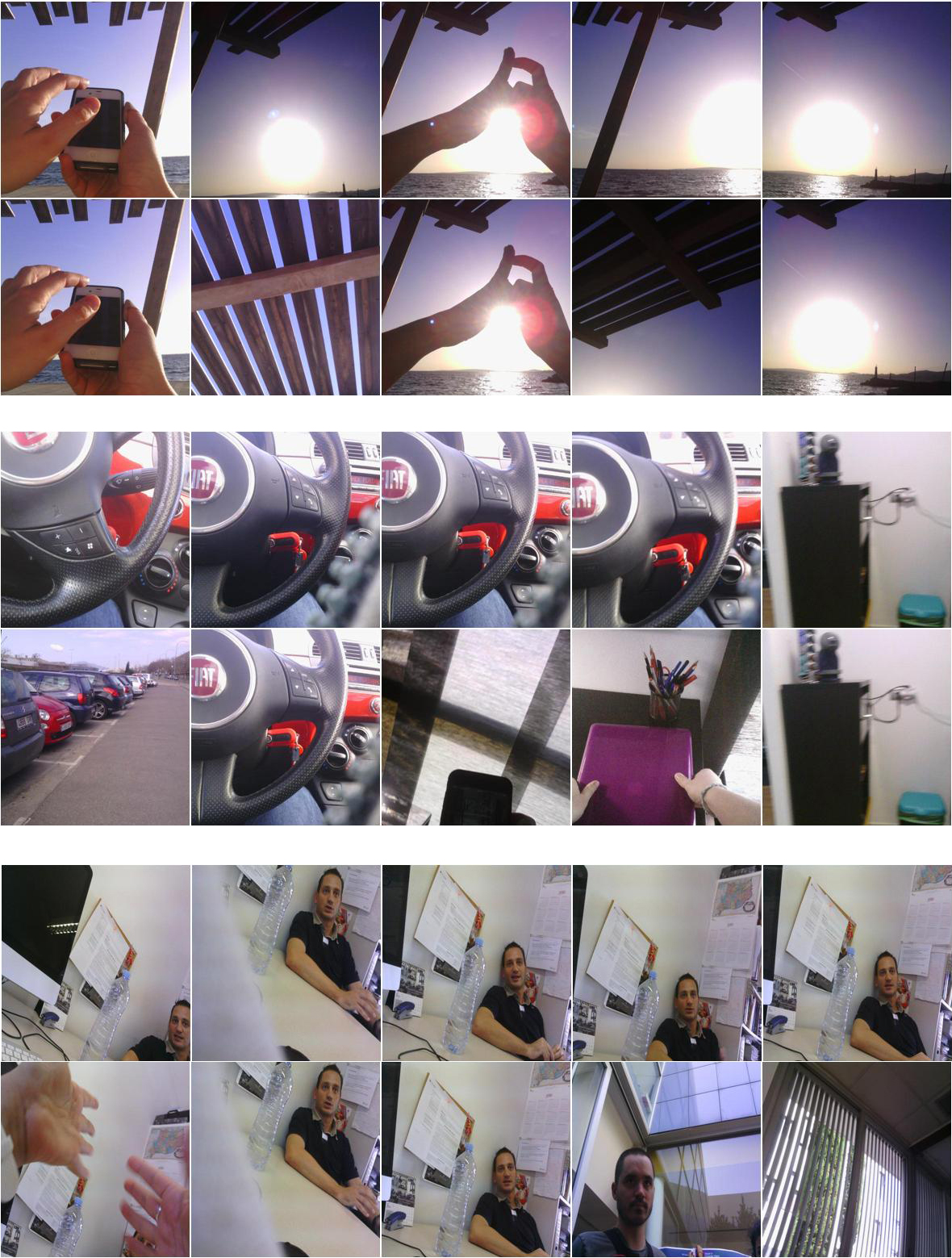}
    \caption{Three examples of the top 5 images obtained before introducing diversity (uneven rows) and after introducing it (even rows).}
    \label{fig:qualitative_diversity}
    \end{center}
\end{figure}

Fig. \ref{fig:qualitative_diversity} illustrates qualitatively the differences obtained when introducing diversity to the ranked list. As we can see, when we introduce the novelty re-ranking step, we are able not only to obtain a more visually acceptable set of images, but also to describe with pictures everything that is happening during the whole event. Thus, we are able to avoid focusing on a single set of high relevant images that picture the same concept, activity or background.

\subsection{Ranking Summary Quality}
\label{subsection:summary_quality}

The first round of blind taste tests posed two questions for each event. Firstly, experts decided whether each presented summary was representative, and secondly, they chose the preferred one 
Given the question \textit{Is the summary representative of the event?}, in 95.74\% the experts agreed with our solution, using the ImageNet features to compute the similarity in the novelty-based re-ranking. 
Using the other alternative, features from Places CNN, the obtained result was 94.33\%.
We conclude that our approach generates representative summavery high portion of events.

The second question asked was: \textit{Which summary do you prefer?} and experts had to choose their preferred summary, allowing just one answer.
The solution based on Imagenet features was chosen in 59.57\% of teh cases, while the one based on Places was selected 53.19\% of the times.
The total adds more than 100\% because some summaries were identical for both configurations.
	
The second round of evaluations aimed to compare our solution based on ImageNet features with a baseline of uniform sampling and and upper-bound defined by the summaries in the ground truth.
Experts were asked to \textit{grade each visual summary from 1 (worse) to 5 (best)}, so we could compute the Mean Opinion Score (MOS) \cite{streijl2016mean} of each solution. 
We adopted MOS as a metric given the highly subjective and complex nature of the task.

\begin{table}[h]
\begin{center}
\caption{Mean Opinion Score for ImageNet, ground-truth and uniform sampling summaries. }
\label{tab:eval2}
\begin{tabular}{ccc}
\hline
\textbf{Our solution}	&\textbf{Ground-truth}	&\textbf{Uniform Sampling}\\ \hline 
\textbf{4,57}	&	  \textbf{4,94}	& \textbf{3,99} \\
\end{tabular}
\end{center}
\end{table}

The performance obtained by uniform sampling (3.99/5) is truly commendable, since this score can be interpreted as a \textit{good} solution. 
The results obtained with our solution are also satisfactory, because they are closer to the ground-truth than to the uniform sampling. 
Experts have shown coherence with their ground-truth giving a 4.94 of Mean Opinion Score.  

%% file: table_informativeness.tex
\begin{table*}[!ht]
\centering
\caption{Best validation accuracy on each set and the respective number of training iterations performed to achieve the results. $SubjX$ represents the anonymized sets for a given Subject.}
\begin{tabular}{|c||c|c|c|c|c|c|c|c|c|c|} \hline 
 & $SubjA_1$ & $SubjA_2$ & $SubjB_1$ & $SubjB_2$ & $SubjB_3$ & $SubjC_1$ & $SubjD_1$ & $SubjE_1$ & $SubjF_1$ & $SubjF_2$\\ \hline \hline 
 \textit{Accuracy} & 0.759 & 0.841 & 0.805 & 0.795 & 0.799 & 0.837 & 0.805 & 0.867 & 0.795 & 0.897 \\ \hline
 \textit{Iteration \#} & 3,600 & 1,000 & 3,600 & 3,400 & 3,800 & 2,600 & 2,200 & 3,600 & 2,400 & 18,000 \\ \hline
\end{tabular}
\label{tab:cross_val_informativeness}
\end{table*}

%% file: 6_conclusions.tex
\section{Conclusions}
\label{sec:conclusions}
In this paper, we presented a novel approach for semantic summarization of egocentric photo stream events, constructed as a ranked list of general semantic criteria.
Our method is based on two main criteria: relevance to optimize semantic diversity in the summary, and novelty to avoid redundancy in the final result. After applying a pre-processing step to filter non-informative images through a new CNN-based method, relevance diversity-aware ranking is obtained by integrating state of the art techniques for saliency detection, object recognition and face detection. This list is re-ranked to reduce redundancy so that each image of the truncated list differs as much as possible from its predecessors. We proposed a new soft metric to rank the informative frames and construct the final summary that does not penalize  summaries equivalent (very similar, but not coinciding exactly) to the ground truth. Experimental results indicate high acceptance and satisfaction of  psychologists achieving mean opinion score of 4.57 out of 5.0. 

%% file: 7_acks.tex
\begin{acks}
This research was supported by contracts SGR1219 and SGR1421 by the Catalan AGAUR office, and TIN2012-38187-C03-01 and TEC2013-43935-R by the Spanish Ministerio de Economia y Competitividad and the European Regional Development Fund (ERDF).
We  acknowledge the support of NVIDIA Corporation for the donation of GPUs.
\end{acks}